\documentclass[acmtog, nonacm]{acmart}
\acmSubmissionID{1197}

\usepackage{booktabs} 

\usepackage{xcolor}
\usepackage{animate}
\usepackage[ruled]{algorithm2e} 
\usepackage{algpseudocode}

\SetAlFnt{\small}
\SetAlCapFnt{\small}
\SetAlCapNameFnt{\small}
\SetAlCapHSkip{0pt}

\acmJournal{TOG}

\begin{document}
\title{Mobius: Text to Seamless Looping Video Generation via Latent Shift}

\author{Xiuli Bi}
\affiliation{
    \institution{Chongqing University of Post and Telecommunications}
    \city{Chongqing}
    \country{China}
    }

\author{Jianfei Yuan}
\affiliation{
    \institution{Chongqing University of Post and Telecommunications}
    \city{Chongqing}
    \country{China}
    }

\author{Bo Liu}
\affiliation{
    \institution{Chongqing University of Post and Telecommunications}
    \city{Chongqing}
    \country{China}
    }

\author{Yong Zhang}
\affiliation{
    \institution{Meituan}
    \city{Shenzhen}
    \country{China}
     }

\author{Xiaodong Cun}
\authornote{Corresponding Author}
\affiliation{
    \institution{GVC Lab, Great Bay University}
    \city{Dongguan}
    \country{China}
    }
\email{cun@gbu.edu.cn}

\author{Chi-Man Pun}
\affiliation{
    \institution{University of Macau}
    \city{Macau}
    \country{China}
    }

\author{Bin Xiao}
\affiliation{
    \institution{Chongqing University of Post and Telecommunications}
    \city{Chongqing}
    \country{China}
    }

\settopmatter{printacmref=false}
\newcommand{\todo}[1]{\textcolor{red}{//TODO// #1}}
\newcommand{\TODO}[1]{\textcolor{red}{//TODO// #1}}
\newcommand{\tocite}{\textcolor{red}{TO Cite}}
\newcommand{\modelname}{LVDM\xspace}

\makeatletter
\DeclareRobustCommand\onedot{\futurelet\@let@token\@onedot}
\def\@onedot{\ifx\@let@token.\else.\null\fi\xspace}

\def\eg{\emph{e.g}\onedot} \def\Eg{\emph{E.g}\onedot}
\def\ie{\emph{i.e}\onedot} \def\Ie{\emph{I.e}\onedot}
\def\cf{\emph{c.f}\onedot} \def\Cf{\emph{C.f}\onedot}
\def\etc{\emph{etc}\onedot} \def\vs{\emph{vs}\onedot}
\def\wrt{w.r.t\onedot} \def\dof{d.o.f\onedot}
\def\etal{\emph{et al}\onedot}
\makeatother

\newcommand{\xiaodong}[1]{{\textcolor{orange}{[xiaodong: #1]}}}
\newcommand{\xd}[1]{{\textcolor{orange}{\textit{#1}}}}

\newcommand{\jianfei}[1]{{\textcolor{blue}{[jianfei: #1]}}}
\newcommand{\jf}[1]{{\textcolor{blue}{\textit{#1}}}}

\begin{abstract}

We present Mobius, a novel method to generate seamlessly looping videos from text descriptions directly without any user annotations, thereby creating new visual materials for the multi-media presentation. 
Our method repurposes the pre-trained video latent diffusion model for generating looping videos from text prompts without any training.  
During inference, we first construct a latent cycle by connecting the starting and ending noise of the videos. Given that the temporal consistency can be maintained by the context of the video diffusion model, we perform multi-frame latent denoising by gradually shifting the first-frame latent to the end in each step. As a result, the denoising context varies in each step while maintaining consistency throughout the inference process.
Moreover, the latent cycle in our method can be of any length. This extends our latent-shifting approach to generate seamless looping videos beyond the scope of the video diffusion model's context.
Unlike previous cinemagraphs, the proposed method does not require an image as appearance, which will restrict the motions of the generated results. Instead, our method can produce more dynamic motion and better visual quality.
We conduct multiple experiments and comparisons to verify the effectiveness of the proposed method, demonstrating its efficacy in different scenarios.
All the code will be made available.
\end{abstract}

\begin{teaserfigure}
\centering
\begin{minipage}{0.49\textwidth}
    \animategraphics[loop,autoplay,width=\textwidth]{8}{figures/gifs/flag/frame_}{0}{51}
    {Prompt: \textit{``A young female activist stands tall, holding a flag high above her head with determination in her eyes. The flag flutters in the breeze, its bold colors contrasting with the backdrop of a city street or public space. Her posture is confident, embodying strength.'' }}
  \end{minipage}
  \hfill
  \begin{minipage}{0.49\textwidth}
    \animategraphics[loop,autoplay,width=\textwidth]{8}{figures/gifs/koala/frame_}{0}{51}
    {Prompt: \textit{``A sleepy koala, nestled comfortably on a tree branch, lazily munches on eucalyptus leaves, its fluffy grey fur blending with the textured bark of the tree. The leaves sway slightly in the breeze as the koala picks them one by one, its black nose twitching with each bite.''}}
  \end{minipage}
\caption{
Without any training, the proposed Mobius can generate seamless looping videos using the pre-trained Text-to-Video latent diffusion model directly. Can you identify the end in the above video? 
\textit{Best viewed with Acrobat Reader. Click the video to play the animation clips. \textbf{We also give these examples in the supplementary video.}} Project page: {\color{red}\url{http://mobius-diffusion.github.io}}.
}
\end{teaserfigure}

\maketitle

\section{Introduction}
\label{sec:intro}

Looping video, also called \textit{cinemagraph} in some research, aims to create a seamless looping video without ends via periodical motions. It is a unique way to share a specific moment's dynamics, which is popular as short videos and animated GIFs on social media, photo-sharing platforms, and screen savers\footnote{\url{cinemagraphs.com}} to create a better user experience. 
However, capturing these looping videos needs huge manual efforts, including the stabilization of the camera, manually annotating the moving object, selecting the animated frames, \etc. 

Previous efforts \cite{10.1111/cgf.12147, holynski2021animating, endless_loops, liao2013automated} make cinemagraphs from the given video or a single image animation. However, due to the difficulty of modeling open-world motion prior, these methods only focus on creating the looping video on the specific kinds, for example, water~\cite{holynski2021animating, mahapatra2023synthesizing, liao2013automated}, periodic pattern~\cite{endless_loops}, portrait~\cite{sadtalker, 10.1111/cgf.12147, bertiche2023blowing}, panoramic~\cite{agarwala2005panoramic, 10.1145/3144455}. 
Since the diffusion model provided universal generative priors for video, current frame interpolation methods~\cite{wang2024generative, loopanimate} can naturally produce the cinemagraph by setting the same beginning and end frames, however, the generated results in frame interpolation will often tend to generate still results in all frames.  
Besides, all current cinemagraph methods focus on simple motions with limited movement, whereas real-world videos are more complex.


We define a new research problem beyond current cinemagraph synthesis which is directly generating the seamless looping video from text description. Different from previous methods which need tricks of a stable camera and only repeating some of the elements, our method aims to generate fully looping videos directly from the pre-trained text-to-video models, which will show more dynamic motions and natural visual effects, including the moving objects, the camera, \etc, by the generative prior. 
This is fully automatic and can generate videos which is unusual in real life. 
However, there are two key challenges in adapting it for our task. First, as the text-to-video diffusion model is trained on natural video, it remains unclear how to adapt it to our looping video generation. On the other hand, the current text-to-video generation model can only generate certain frames during inference. However, a short video might not provide a good representation of real-world dynamics. 

Thus, we present Mobius to solve these problems in a training-free manner, where the key observation is that each frame should be considered equally important in the video final video. To this end, firstly, we propose a latent shift strategy in denosing. We construct a cycle utilizing all the noisy latents from the first frame to the end frame. Then, we shift its position by adding the first frame to the last to build the new noisy latent for denoising.
Thus, the video model maintains temporal consistency in each denoising step, and each video is equally considered.
For the generation in the longer context, the proposed latent shifting strategy naturally enables longer looping video generation by a longer denoising sequence cycle. 
However, if we directly generate the longer video utilizing this method, the generated results are also influenced by the inaccurate position embedding and frame-variant 3D VAE. Thus, we extend the rotary position embedding by an NTK-aware interpolation method inspired by the long context Large Language Model~\cite{ntk_aware} and propose a frame-invariance method for latent decoding.
Based on these modifications, the proposed method can directly utilize the pre-trained video diffusion model to generate high-quality cinemagraphs from text descriptions. Besides, we also show that the proposed latent shift can also work well for longer video generation tasks. Finally, the experiments demonstrate the qualitative and quantitative advantages of our approach. 

Overall, the contribution can be summarized as:
\begin{itemize}
    \item We conduct a new research problem for the open-domain seamless looping video generation from text description using a pre-trained text-to-video diffusion model.
    \item We propose a latent shifting strategy to interactively denoise the latent in each step so that we can generate the looping video and it can be in arbitrary lengths.
    \item The detailed experiments show that the proposed method can achieve state-of-the-art performance on looping video generation and we also give the applications on longer video generation.
\end{itemize}

\section{Related Work}

\subsection{Cinemagraphs and Looped Video Generation}
Our task is similar to cinemagraphs, which aim to produce looping videos by manipulating an input video manually. However, the manual creation of cinemagraphs is a time-consuming process, even for professional artists.
Previously, learning-based methods faced difficulties in generating or editing an entire video. As a result, prior techniques only applied to specific patterns to create cinemagraphs, for example, water~\cite{holynski2021animating, mahapatra2023synthesizing, liao2013automated}, periodic pattern~\cite{endless_loops}, portrait~\cite{sadtalker, 10.1111/cgf.12147, bertiche2023blowing}, panoramic~\cite{agarwala2005panoramic, 10.1145/3144455}. As for representative work,
Endless Loops~\cite{endless_loops} utilizes CRF to compute loop shifts, and it can only work on the repeated pattern.  \cite{holynski2021animating} presents an image animation method to generate the moving water from a single image utilizing Eulerian motion fields. Text-to-cinemagraphs~\cite{mahapatra2023synthesizing} further extend it by the pre-trained text-to-image stable diffusion model. Several methods ~\cite{generative_image_dynamic, niu2024mofa, shi2024motion} present a two-stage framework to generate the video with trajectory control. However, they only work on certain object types or need manual trajectory design.
While LoopAnimate~\cite{loopanimate} employs multi-stage training and symmetric guidance to achieve looped generation, their generated results are too still.
Besides, we can naively utilize the generative frame interpolation methods~\cite{wang2024generative,xing2024tooncrafter} based on the video diffusion model for a generation by setting the same start and end keyframes. However, since the original frame interpolation model is not trained for cinemagraph, the generated results might also be still. Besides, these methods involve additional larger-scale training for generation, which might cause forgetting problems.
Differently, we directly generate the looped video from the text description, yet with better visual effects, such as the whole movement of the camera and object motion.

\begin{figure*}[t]
\centering
\includegraphics[width=\textwidth]{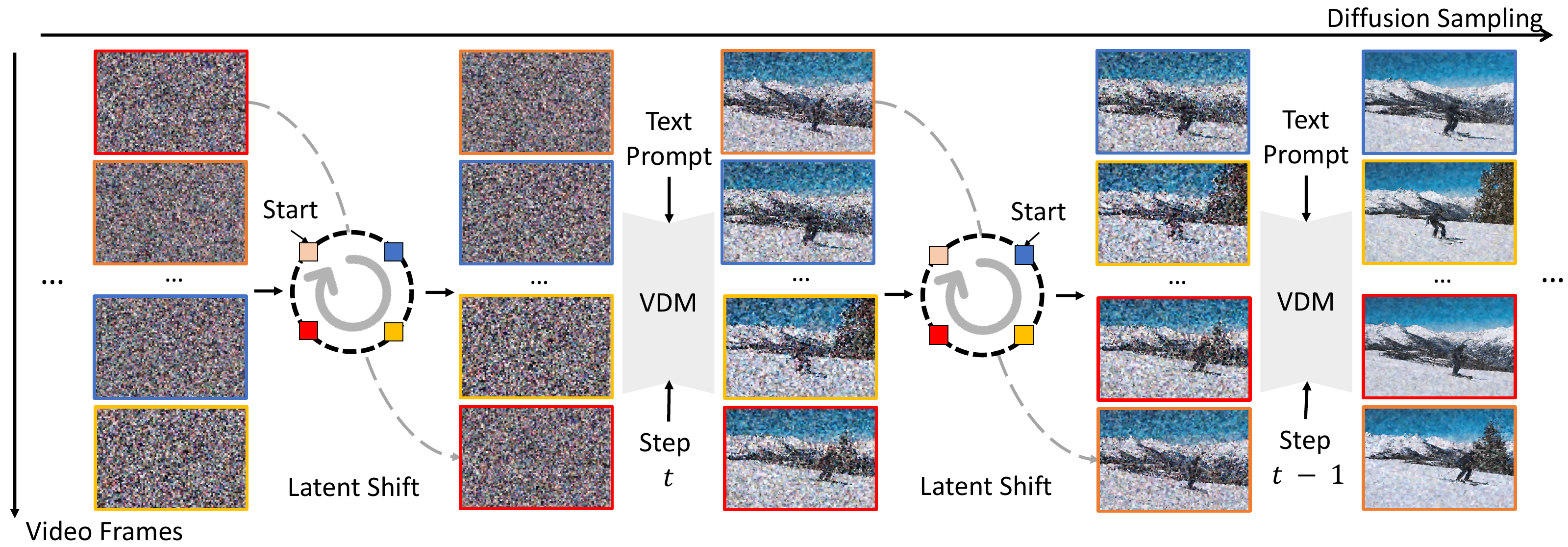}
\vspace{-2em}
\caption{\textbf{Latent Shift for looping video generation}. Taking 4 latent toys pre-trained Video Diffusion Models~(VDM) as an example, we build a latent cycle and shift the start point in each denoising step in inference for text-guided looping video generation. Notice that, the shifting is conducted in the latent space, we emit the latent encoder and decoder for easy understanding.}
\vspace{-1em}
\label{fig:PLS}
\end{figure*}

\subsection{Video Generation in Diffusion Model Era}
Due to the stabilizing training process of the Diffusion Model~\cite{ddim, ddpm}, video generation has had a big breakthrough in recent years. Eary works~\cite{imagen-video, make-a-video} directly generate high-resolution videos from cascade models of spatial and temporal layers directly in pixel space.
On the other hand, utilizing the pre-trained text-to-image models~\cite{ldm}, \ie, Stable Diffusion, as the base model, many works try to add additional layers to keep temporal consistency. \cite{videocrafter1, modelscope, magic-video} add temporal attention modules to the base model and train in an end-to-end fashion. Besides, \cite{animatediff} finds that training the models by temporal layers only has a better visual quality. \cite{videocrafter2} proposes a method to increase the visual qualities by a two-stage image and video joint training process. However, these methods only create a short video with limited motions, which restricts its applications in real-world cases. 
Besides the text-to-video diffusion model, new works also train image-to-video models for generation, which is also related to our task. For example, Stable Video Diffusion~\cite{svd} fine-tunes the text-to-video diffusion model with a high-quality data pipeline. DynamicCrafter~\cite{xing2025dynamicrafter} shares a similar idea and trains on the video diffusion model. ToonCrafter \cite{xing2024tooncrafter} and Generative image in-between \cite{wang2024generative} are further finetuning the image-to-video models for the generative frame interpolation. However, as we discussed before, directing utilizing the frame interpolation methods for our task might have issues with the too-still motion.
Recently, Sora \cite{sora} has made a big step in video generation via denoising transformers~(DiT~\cite{DiT}), showing the scalability and advantages. 
Thus, the more recent video generation methods~\cite{mochi, cogvideox,hunyuanvideo} are based on the DiT structure, which has better motion and temporal consistency than previous methods.  

Besides, since these pre-trained large diffusion models are trained from larger-scale datasets, we can repurpose these models for the new task without training. For instance, in the field of image/video editing, works such as Prompt to Prompt~\cite{prompt-to-prompt}, FateZero \cite{fatezero}, and MasaCtrl~\cite{masactrl} have achieved zero-shot editing through attention control. Meanwhile, there also contains some methods that have provided foundational discoveries for zero-shot editing~\cite{yu2023animatezero, freedom} and improving the performance without additional training~\cite{freeinit, freeu}. In this paper, we utilize the most popular open-sourced DiT-based video generation model, \ie, CogVideoX~\cite{cogvideox}, as the base model for looped video generation in a training-free manner. 


\subsection{Longer Video Generation in Diffusion Model}
\label{longer_video_generation}
Our looping videos can be considered an infinitely longer video generation. In current methods, due to the limited latent length in training the pre-trained text-to-video generation models, several methods are proposed to modify the denoising process of the original diffusion model for new purposes. For the longer video generation, Gen-L-Video~\cite{gen-l-video} uses the weighted sum of different short latent segments in the overlapping area to alleviate the inter-frame continuity issue. However, this method significantly increases the inference time and can lead to smooth transitions between frames. 
FreeNoise~\cite{freenoise} introduces a shuffled latent sequence design and uses attention-based weighting to maintain visual consistency in long videos. However, since the latent changes only occur in the shuffling, the resulting video motion may appear too static and is prone to out-of-memory~(OOM) errors. FIFO~\cite{fifo} uses diagonal denoising for long video generation, maintaining the consistency and coherence of the video. However, there is a training inference gap for reasoning at different noise levels, and it lacks global information modeling. 
Video-Infinity~\cite{video-infinity} uses distributed inference to facilitate global and local information interaction, achieving video consistency while accelerating inference. However, an important limitation is the need for multiple GPUs to run simultaneously, and the quality of generating longer videos is not very good. 
DiTCtrl~\cite{ditctrl} utilizes a mask-based attention-sharing mechanism to maintain semantics, as well as a latent mixing strategy to achieve smooth transitions between video frames. However, this also brings about a significant amount of additional computational costs. 
These longer video generation methods change the combination of latent in the test time to control the generated content in the diffusion process, which inspired our looping video generation from text directly. 

\section{Method}


Given the text prompt, we design a training-free method for generating the looping video by shifting the noise in each inference step of the pre-trained video diffusion model, so that all the frames will be considered equally in the final generated video.
Below, we first introduce the basic paradigm of text-to-video to better understand our method in Sec.~\ref{sec:preliminary}. 
Then, we introduce the proposed \textit{Latent Shifting}, which iteratively transforms the position of the latent in each step~(Sec.~\ref{sec:PLS}). Since directly utilizing the looping latents will show artifacts when 3D VAE decoding, we design a frame-invariant decoding to decode looping video~(Sec.~\ref{sec:fivae}). Finally, we introduce the \textit{Rotary Position Encoding interpolation} to model global positional information when generating the longer looping videos in Sec.~\ref{sec:DyRoPE} and give some applications in Sec.~\ref{sec:application}, respectively.

\subsection{Preliminary: Text-to-Video Latent Diffusion Model}
\label{sec:preliminary}
Taking one specific video diffusion model, \ie, CogVideoX~\cite{cogvideox}, as an example, we introduce the basic concepts and knowledge of the text-to-video latent diffusion model. 
Current large text-to-video models are all based on the latent diffusion model~\cite{ldm}. The latent diffusion model contains an auto-encoder $\mathcal{E}(\cdot)$, $\mathcal{D}(\cdot)$ for compressing the videos into the latent space. In the most advanced video diffusion models~\cite{sora, cogvideox, mochi}, the compression of videos in both the spatial and temporal domains represents the crucial factor for realizing better visual and temporal qualities. Then, following the Denoising Diffusion Probabilistic Models~\cite{ddpm}, for training, the input $F$ frame video clip $ \mathbf{\upsilon} \in \mathbb{R}^{F \times H \times W \times 3} $ with width $W$ and height $H$ is first converted to the latent space $\mathbf{z}_0$, where $\mathbf{z}_0 = \mathcal{E}(\mathbf{\upsilon}) = [z_0^1; ... ; z_0^f] \in \mathbb{R}^{f \times h \times w \times c}$. $h,w,f$ are the compressed height, width, and frame in the latent space, respectively. Then, the latent diffusion model $\epsilon_\theta$ is trained to denoise its perturbed version $\mathbf{z}_{t}$. For noise $\epsilon \sim \mathcal{N}(0, \mathbf{I})$, the time step of diffusion model $t \sim \mathcal{U}([1,...,T])$, the text prompt $c$, this denoising diffusion model is trained to minimize the following loss:
\begin{equation}
    \mathcal{L}=||\epsilon-\epsilon_{\theta}(\mathbf{z}_{t};c,t)||_{2}^{2}.
\end{equation}
Here, the denoising network $\epsilon_{\theta}$ is based on the DiT~\cite{DiT} architecture. 

After training, giving any noise latents $[z_{t}^{1};...;z_{t}^f] \sim \mathcal{N}(0, \mathbf{I})$ for video generation, and a diffusion sampler $\Phi(\cdot)$, such as DDIM sampler~\cite{ddim}, the diffusion model generate the final clear video via an $T$-step iterative denoising, where $t$-th denoising step is expressed as:
\begin{equation}
    [z_{t-1}^{1}; ...; z_{t-1}^{f}] = \Phi([z_t^1; ...; z_t^f], t, c; \epsilon_{\theta} ),
\end{equation}
where $z_{t}^{i}$ denotes the latent of $i$-th frame at time step $t$. Notice that, the context length of the video diffusion model is restricted by the denoising network $\epsilon_{\theta}$, and each latent has the unchanged position when inference.

Finally, we could generate a video by the pre-trained latent decoder $\mathcal{D}(\cdot)$ of the 3D VAE as: 
$\mathbf{\upsilon}' = \mathcal{D}(\mathbf{z}_{0}')$. Notice that, since the 3D VAE of the video diffusion model supports both image and video generation, they usually treat the first frame differently in temporal compression.

\subsection{Latent Shifting}
\label{sec:PLS}

The text-to-video diffusion model is trained on a multi-frame latent diffusion model, where multiple latents are sent into the denoising network for a generation. 
Since our looping video requires each frame to be considered as the first frame, we thus need to make each latent have the temporal consistency of the previous latent and the end latent. As shown in Figure.~\ref{fig:PLS}, we first build a cycle latent list for denoising by connecting the first frame latent and the last. Then, for each denoising step, we shift the first frame to the last to build a new multi-frame latents for generation. After multi-step denoising, we can maintain the whole temporal consistency of the entire video. 

Formally, given the inference context $f$ of a video diffusion model, we can generate the looping video which contains $N$ latents, where $N = n \times f$ and $n$ are the multiple factors for longer looping video generation. Firstly, we initialize all the latent as $[z_{T}^{1};...;z_{T}^N] \sim \mathcal{N}(0, \mathbf{I})$, then, for $t$-th denoising step, we shift the start point of the denoising context by $j = (t \times s) \mod N$, where $s$ is the skip step of each iteration.  Since we also need to maintain the $f$-frame inference restriction in the pre-trained diffusion model, the denoising step of this step can be formulated as:
\begin{equation}
    [z_{t-1}^{j}; ...; z_{t-1}^{j+f-1}] = \Phi([z_t^{j}; ...; z_t^{j+f-1}], t, c; \epsilon_{\theta} ),
\end{equation}
where $\Phi$ is a DDIM Sampler~\cite{ddim} as introduced before. When $j + f -1 > N $, our cycle list creates the denoising latents by concat the $[z_t^{j},...,z_t^{N}]$ and $[z_t^{1},...,z_t^{f-(N-j+1)}]$.

Our latent shifting algorithm utilizes the multi-frame denoising steps in the diffusion model and the temporal consistency denoising of the video diffusion model for looping video generation. Notice that, since this latent denoising method can be any length, our method can produce any length inference looping videos and can also be utilized in the longer video generation.

\begin{figure}[!t]
\centering
\includegraphics[width=\columnwidth]{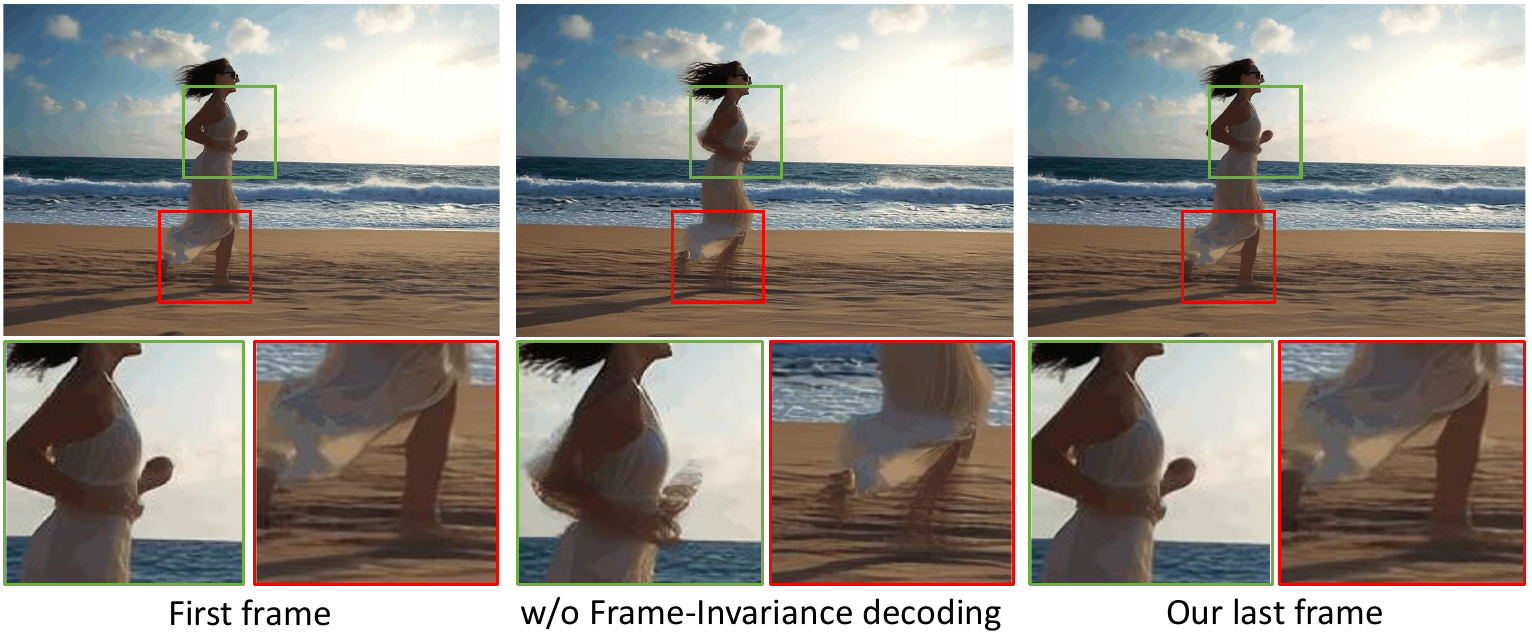}
\vspace{-2em}
\caption{Frame-invariance latent decoding reduces the artifacts caused by the 3D VAE decoding.}
\vspace{-1em}
\label{fig:vae}
\end{figure}

\subsection{Frame-Invariance Latent Decoding}
\label{sec:fivae}
To meet the demands of both text-to-video and text-to-image joint training, the latent compression of current state-of-the-art video generation models~\cite{cogvideox} does not compress each frame equally in the temporal dimension. In detail, CogVideoX employs a 3D VAE structure that compresses video frames both in spatial and temporal compression. 
However, the first latent frame employs a special encoding and does not do any compressions, while subsequent frames are encoded with the standard $4 \times$ compression for the motion similarity. 
In latent decoding, it utilizes the first three latent to generate the first night frames of video.
This inconsistent treatment of latent is inherently incompatible with our proposed latent shift method for looping video generation since we aim to produce a looping video in which each frame should be considered equally. If we directly utilize the original 3DVAE, it results in artifacts in the generated first frame due to the $4\times$ compression, as shown in Figure~\ref{fig:ablation}.
To mitigate this issue, we copy the last three latents and insert them before the first latent as redundant frames to counteract the special compression of the first frame. Then, in the generated video, we remove the redundant generated frames by the added latent.


\subsection{Rotary Position Embedding Interpolation}
\label{sec:DyRoPE}

CogVideoX~\cite{cogvideox} uses Rotary Position Embedding~(RoPE) to give positions in the attention model for denoising, which aims to achieve relative position encoding via absolute rotary position. However, if we directly utilize the original PoPE for our longer looping video generation task, the longer context does not match the original text-to-image model. 
To address this issue, we utilize a RoPE~\cite{rope} interpolation method for globally latent coding in the temporal dimension, inspired by the NTK-Aware interpolation in the longer context large language model~\cite{ntk_aware}. 

\begin{figure}[!t]
\centering
\includegraphics[width=0.8\columnwidth]{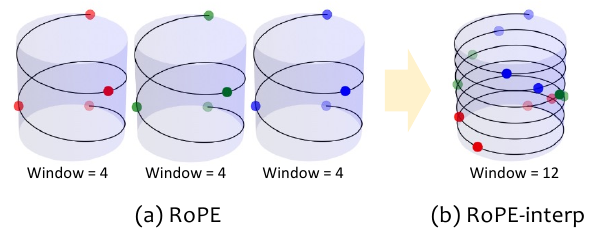}
\vspace{-1em}
\caption{We illustrate this with the example of the toy latent video diffusion model with a context window equal to 4. The utilized RoPE-Interp. enables longer video context without training by interpolation.}
\label{fig:DyRoPE}
\end{figure}

Given the query vector at the $m$ position $q_{m}$ and the key vector at the $n$ position $k_{n}$ in the attention, RoPE introduces absolute positional information before calculating attention as follows:
\begin{equation}
\label{eq:qk_rope}
\begin{aligned}
    Q_{m}&=RoPE(q_{m},m)=q_{m}e^{im\theta},\\
    K_{n}&=RoPE(k_{n},n)=k_{n}e^{in\theta}.
\end{aligned}
\end{equation}
\\
Here, $\theta=\operatorname{diag}(\theta_{0},\cdots,\theta_{d/2-1})$ is a pre-define diagonal matrix, where $\theta_i=b^{-2(b-1)/d}$, with $b=10000$, and $d$ represents the vector dimension. Then, we perform an inner product calculation to obtain the attention weights $A_{m,n}$ as follows:
\begin{equation}
\label{eq:qk_product}
\begin{aligned}
    A_{m,n}=\mathrm{Re}[\left< Q_{m},K_{n} \right>]=\mathrm{Re}[\left< q_{m},k_{n} \right>e^{i(m-n)\theta}].
\end{aligned}
\end{equation}
The result can be transformed into a value related to $m-n$, thus achieving relative position encoding.

To extend the encoding for longer lengths, we scale the base $b$ as follows: 
\begin{equation}
\label{eq:b_base_dim}
\begin{aligned}
    b^{\prime}=b\cdot k^{d/(d-2)}
\end{aligned}
\end{equation}
Here, $b^{\prime}$ denotes the result after scaling, $k$ represents the multiple by which the video length increases, and $d$ indicates the dimension of the latents vector. Fig.~\ref{fig:DyRoPE} gives an illustration on how the RoPE-Interp. works. Since our core idea is to make each frame equal in the generation, we also try two different schemes to add the RoPE-Interp. to the features. The first one is the \textit{shifted RoPE-Interp.}, where RoPE changes along with the latents, and another is the \textit{fixed RoPE-Interp.}, where RoPE remains unchanged while the latents shift. We provide a more detailed comparison in the experiments. 

\subsection{More Application: Longer Video Generation}
\label{sec:application}
Longer video generation is an active research topic in current video generation since current text-to-video generation methods can only generate videos with limited context. The proposed latent shift naturally supports longer video inference beyond the training context by a non-cycle latent displacement. We utilize the same RoPE interpolation as we introduced before to correct the position of the latent. We give some examples in the supplemental videos.

\section{Experiment}
\subsection{Settings and Implement Details}

\paragraph{Implementation details}
Our method is based on the pre-trained state-of-the-art open-source latent video diffusion model, CogVideoX-5B~\cite{cogvideox}. Notice that, we only modify the latent input of the diffusion model, our method might also work on any newly designed text-to-video latent diffusion models~\cite{mochi, hunyuanvideo}, without training.
Each video has a resolution of 480x720, and the inference step is set to 50 following a standard DDIM sampling strategy. Other parameters are the same as the default settings of CogVideoX. To evaluate the proposed methods, we choose 140 prompts from VBench \cite{vbench} and EvalCrafter~\cite{evalcrafter} and use GPT~\cite{chatgpt} to expand them into more detailed descriptions. All the experiments are conducted on a single NVIDIA H100 GPU. Since we only add a temporal latent shift in each step denoising, the proposed method has a similar inference speed compared with direct generation.

\paragraph{Baseline}
Since there is no previous work for open-domain looping video generation from a text description, we majorly compare two generative interpolation methods and one method from the community. The first generative interpolation method is \textit{Svd-Interp.} from Generative Image Inbetween~\cite{wang2024generative}, which is trained on the stable video diffusion model~\cite{svd} for frame interpolation. The other generative interpolation is \textit{CogX-Interp.}\footnote{\url{https://github.com/feizc/CogvideX-Interpolation}}, which is also trained from the image-to-video model of the CogVideoX for frame interpolation. 
To compare, we consider the first frame of our generated results for the starting and ending key frames of the interpolation. Notice that these two methods are based on larger-scale training for frame interpolation. Our method generates the looping video from the text description directly.
\textit{Latent Mix} is a method to achieve this looping video, which has been reported on Github\footnote{https://github.com/THUDM/CogVideo/issues/149}, we compare with this method directly.

\begin{table}[]
\caption{Quantitative experimental results for different methods under the numerical evaluation metrics. * for the interpolation-based method, we utilize our generated first frame for the start and end keyframe, thus the MSE between the two frames is the oracle value. }
\vspace{-1em}
\begin{tabular}{lccccc}
\toprule
\multicolumn{1}{c}{\textbf{}} & MSE$\downarrow$ & FVD$\downarrow$           & \begin{tabular}[c]{@{}c@{}}CLIP\end{tabular}$\uparrow$ & \begin{tabular}[c]{@{}c@{}}Motion \\ Smooth\end{tabular}$\uparrow$ & \begin{tabular}[c]{@{}c@{}}Dynamic\\ Score\end{tabular}$\uparrow$ \\ \hline
Svd-Interp.*    & {\color{gray}{18.30}}    &5.66     &32.08     &0.9950       & 0.0667             \\
CogX-Interp.*  & {\color{gray}{15.59}}    &28.60    &31.88     &0.9830       &0.3333        \\ \hline
CogVideoX      &66.89    &56.02    &32.19     &0.9738       &0.7056             \\
Latent Mix     &45.17    &60.02    &31.99     &0.9749       &\textbf{0.7273}          \\ 
Ours           &\textbf{25.43}    &\textbf{40.78}    &\textbf{32.24}     &\textbf{0.9850}       &0.4722            \\ \bottomrule
\end{tabular}
\vspace{-1em}
\label{tab:looping}
\end{table}

\paragraph{Evaluation Metrics}
We report the MSE of the first frame and the last frame in the generated videos due to the looping video has the same first frame and the end frame. For the overall video quality, we utilize the widely used FVD~\cite{fvd} and CLIP Score~\cite{clip-score} for comparison. Besides, we give the overall video smoothness and dynamic score of the whole video from VBench~\cite{vbench}.

\begin{figure*}[t]
\centering
\includegraphics[width=\textwidth]{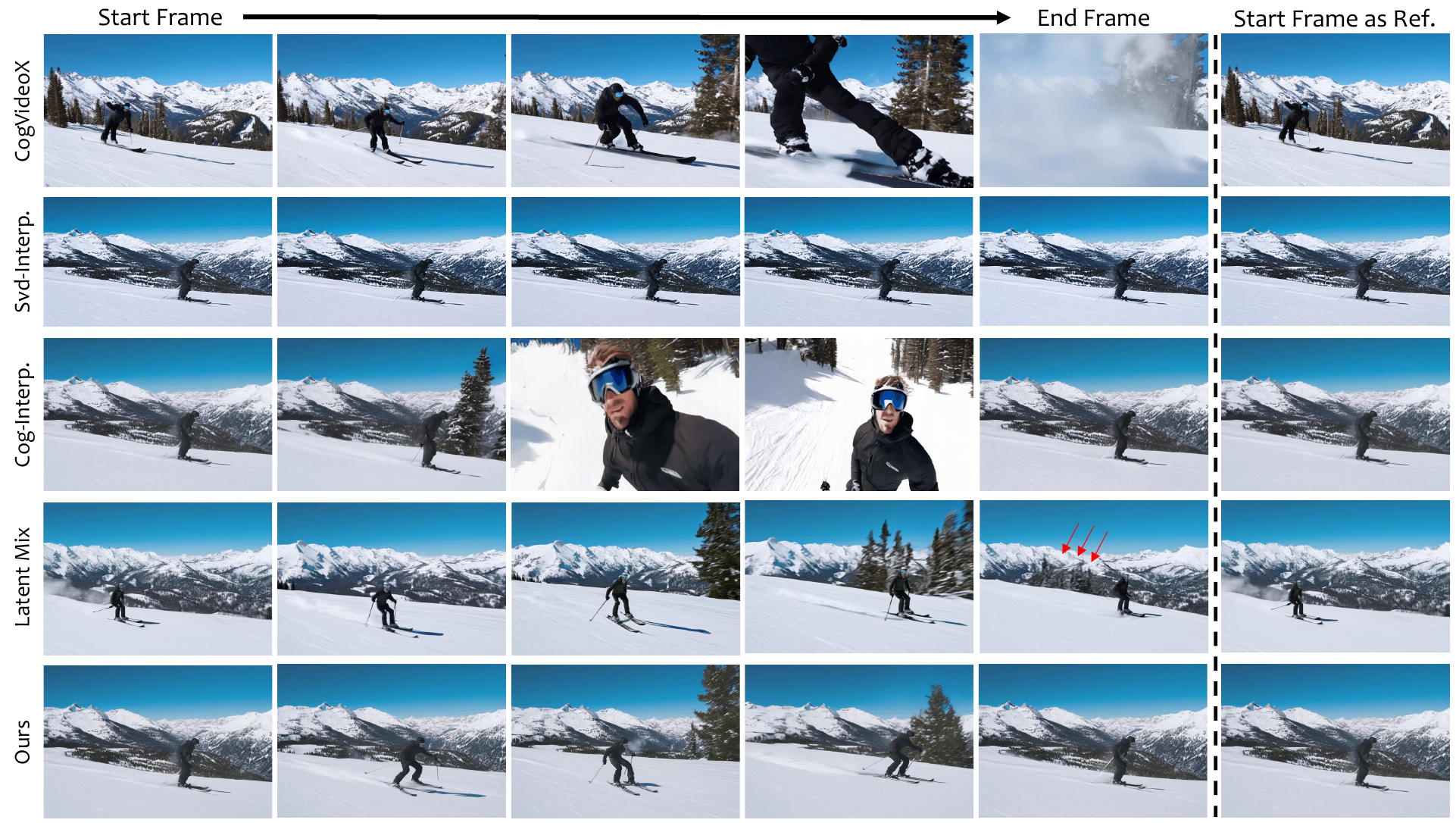}
\vspace{-2em}
\caption{Compare with other methods. We give the first frame, the intermediate frame, and the last frame for comparison. Notice that, both Svd-Interp. and Cog-Interp. are frame-interpolation methods, we manually give the same start frame and end frame as key-frames.}
\label{fig:single_compare}
\end{figure*}

\subsection{Comparison with Other Methods}
\begin{figure*}[t]
\centering
\includegraphics[width=\textwidth]{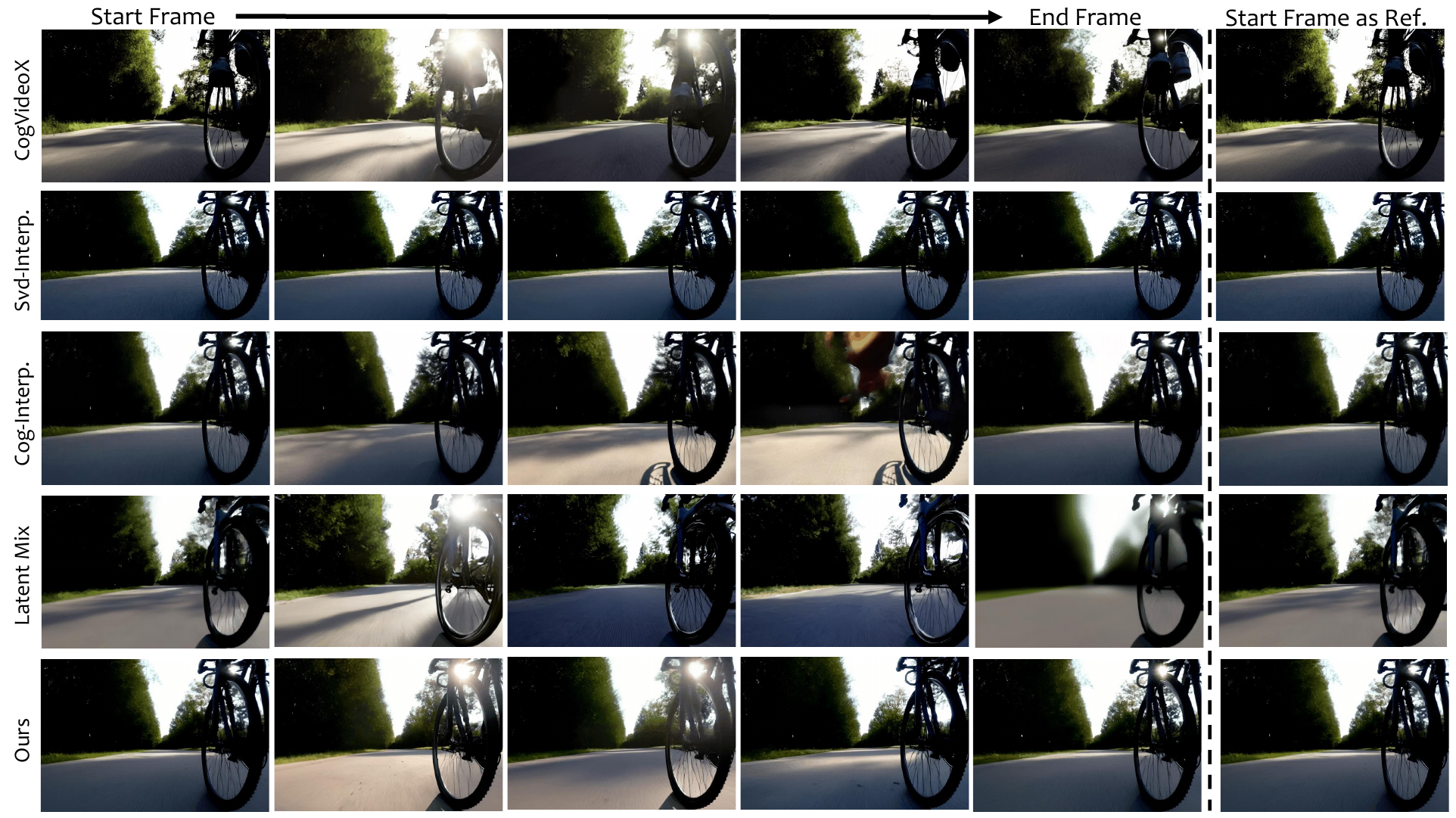}
\includegraphics[width=\textwidth]{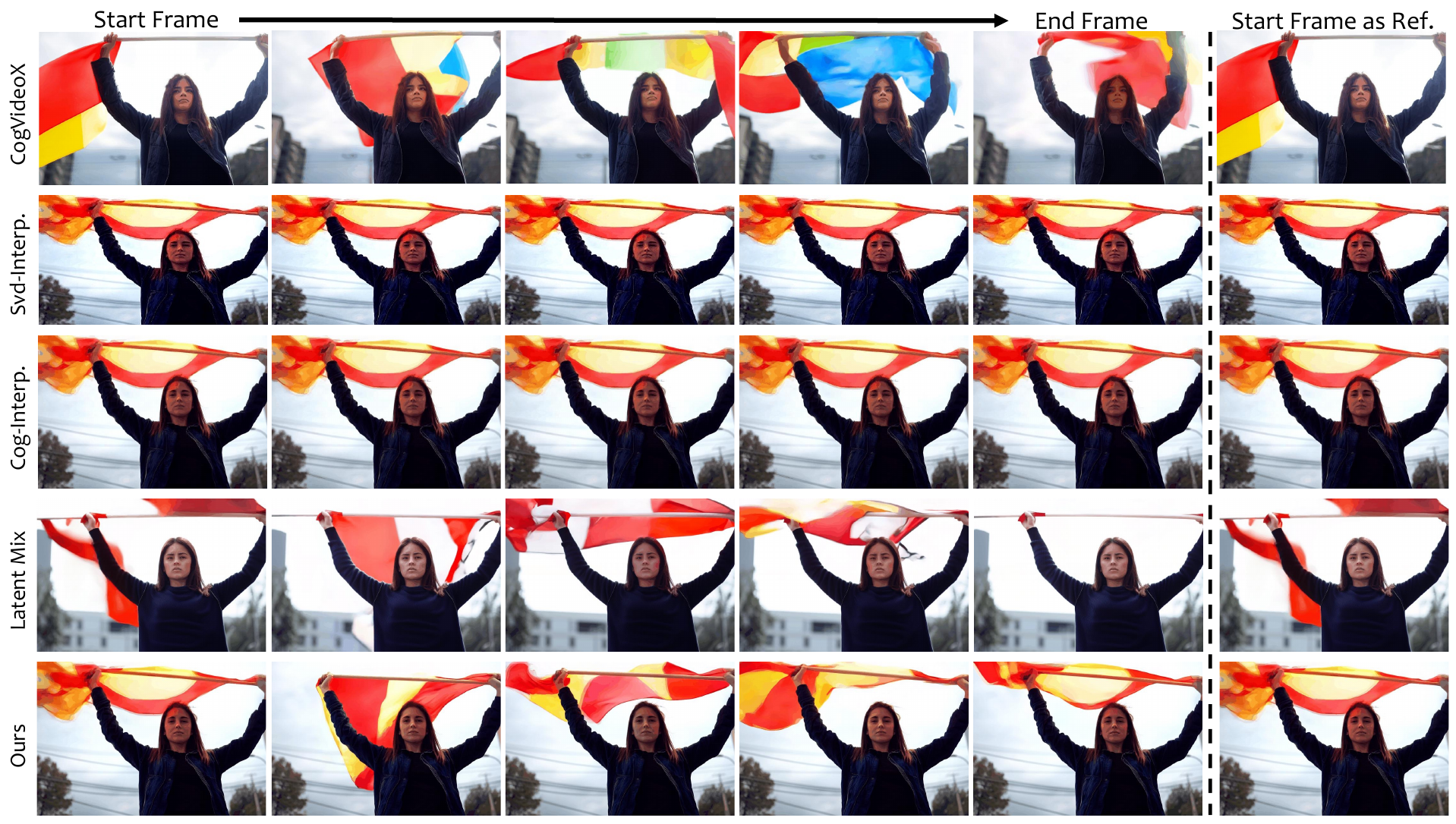}
\caption{More comparisons on the looping video generation.}
\label{fig:additional}
\end{figure*}

As introduced before, since current cinemagraph methods can not work on open-domain looping video generation, we compare our method with the state-of-the-art generative frame interpolation methods introduced in the baseline section. As shown in Fig.~\ref{fig:single_compare}, the baseline interpolation methods may produce still results or generate content that is far away from the start frame and the end frame. The latent mix method blending the initial and final latent may result in artifacts in the end frame. Differently, the proposed method can generate the same start and end frames without noticeable differences. Due to the page limitation, we give more examples in Fig.~\ref{fig:additional} and the supplementary video.

As for the numerical comparison, as shown in Tab.~\ref{tab:looping}, the proposed method shows a better visual quality and text-video alignment than previous methods. Besides, we also achieve a relatively higher score with both motion smoothness, video dynamic, and the MSE between the first frame and the last frame, which shows the advantage of the proposed methods. We argue that although the latent mix method gives much dynamic video, the generated content might not be a looping one according to the MSE between the first and the last frame. 
Evaluating the looping videos using current automatic evaluation metrics is also difficult, so we conduct a subjective user study to prove the proposed method's effectiveness further. 
In detail, we invite 23 participants to rank ten questions across three aspects, totaling 690 opinions under five different methods. Each participant will be asked to rank the overall visual quality of the video, the consistency of the video frames, and how dynamic the video is, on a scale of 5 to 1. Finally, we calculate the average score of these opinions. As shown in Table~\ref{tab:user_study}, our method outperforms others in visual quality, temporal quality, and video dynamic.


\begin{table}[t]
\caption{User Study Results.}
\vspace{-1em}
\begin{tabular}{lccc}
\toprule
\multicolumn{1}{c}{} 
&\begin{tabular}[c]{@{}c@{}} Temporal \\ Consistency\end{tabular}$\uparrow$ 
&\begin{tabular}[c]{@{}c@{}} Visual \\ Quality \end{tabular}$\uparrow$  
& \begin{tabular}[c]{@{}c@{}} Video \\ Dynamic \end{tabular}$\uparrow$   \\ \midrule
CogVideoX      &3.34     &3.62     & 3.68      \\
Svd-Interp.    &1.63     &1.71     &1.53            \\
CogX-Interp.   &2.22     &2.08     &2.17    \\
Latent Mix     &3.52     &3.44     &3.52      \\ 
\midrule
Ours           &\textbf{4.30}     &\textbf{4.15}     &\textbf{4.10}     \\ \bottomrule
\end{tabular}
\label{tab:user_study}
\end{table}

\subsection{Ablation Studies}
\begin{figure}[!t]
\centering
\includegraphics[width=\columnwidth]{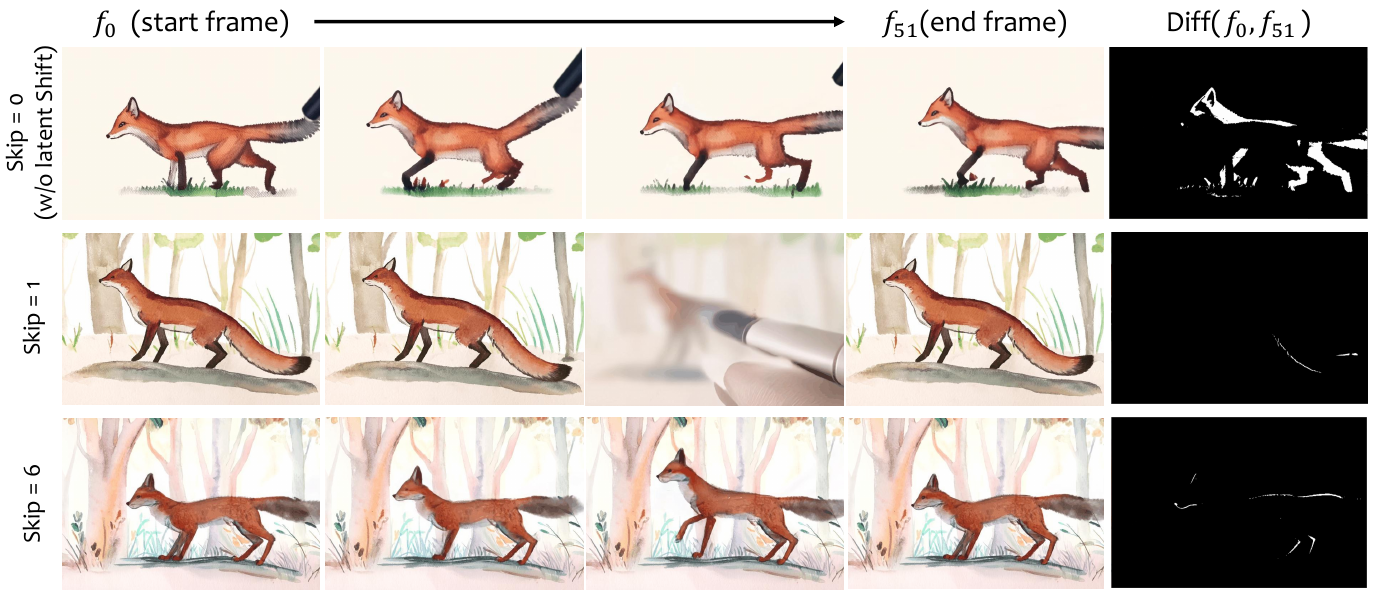}
\vspace{-2em}
\caption{\textbf{Ablation study on different latent skip.} The shift step in each denoising iteration will also influence the generated content.  }
\label{fig:ablation}
\end{figure}

We have given the example in Fig.~\ref{fig:vae} to validate the effectiveness of the proposed frame-invariance latent decoding. Here, we give more ablation studies on the ROPE-Interp. and the skip step in our latent shifting. When performing a latent shift, we can shift the latent $s$ step for denoising, where a small step will be similar to the original inference. As shown in Fig.~\ref{fig:ablation}, when shifting the latent 6 steps in each denoising, the generated content is in a balance of the generated content and the motion. Differently, a small skip will show obversely artifacts.

\begin{figure}[!t]
\centering
\includegraphics[width=\columnwidth]{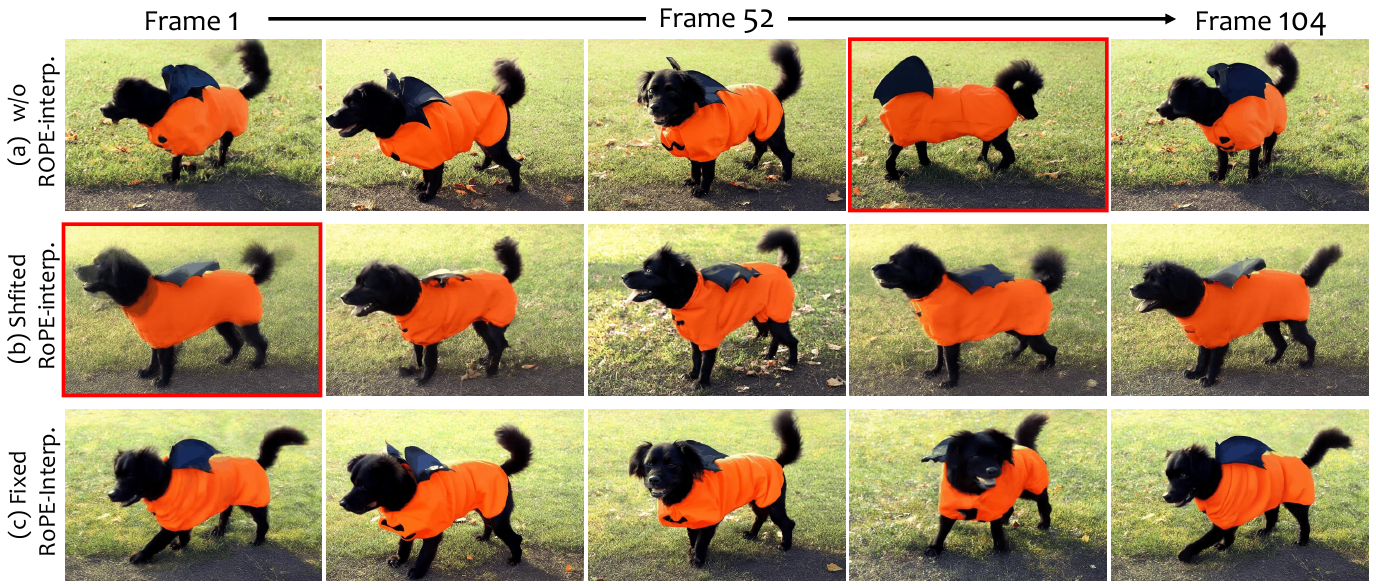}
\vspace{-2em}
\caption{\textbf{Ablation study on RoPE-Interp.} Under the implementation of latent shifting, different RoPE strategies can have a significant impact on the content of video generation.}
\label{fig:rope}
\end{figure}

We also conduct experiments on the RoPE interpolation. In the method, we give two different ways to utilize the interpolated RoPE. As shown in Fig.~\ref{fig:rope}, the fixed RoPE-Interp performs well in our longer video looping generation, allowing each frame to be treated as the first frame during video generation, thereby achieving better looping results.

\subsection{Applications on Longer Video Generation}



Since the proposed latent shifting can naturally work for longer video generation, we also compare our method on longer video generation, where these baselines have been introduced in Section ~\ref{longer_video_generation} for details. 
As shown in Figure~\ref{fig:single_compare}, directly increasing the size of the latent causes video quality collapse. \textit{Gen-L-Video}~\cite{gen-l-video} produces overly smooth transitions in the background and excessive changes in the direction of the seagull. \textit{FreeNoise}~\cite{freenoise} tends to keep the seagull's orientation constant, the static nature of the image caused by latent shuffling is immediately apparent, and the phenomenon of the seagull having three legs also occurs. Although \textit{FIFO}~\cite{fifo} achieves better motion changes and video coherence, the issues of the seagull changing direction twice in a row and having three legs persist. \textit{DiTCtrl}~\cite{ditctrl} improves the seagull's orientation issue, but still has problems with the defective generation of the seagull's head in the first frame and the three-legged issue. In contrast, the proposed method maintains the seagull's orientation while ensuring coherent video motion. It does not exhibit the issue of the seagull having three legs, thereby achieving superior long video generation.
We give the full comparison in the supplementary video. As for the numerical comparison, we conduct the experiments on the same prompts of our looping video generation and calculate the main numerical results in Tab.~\ref{tab:single} utilizing the well-known metrics from previous studies~\cite{ditctrl,freenoise}.

\begin{table}[]
\caption{Comparing with other longer video generation methods.}
\vspace{-1em}
\begin{tabular}{lccc}
\toprule
\multicolumn{1}{c}{\textbf{}} &  FVD$\downarrow$          & \begin{tabular}[c]{@{}c@{}}CLIP \\ Score\end{tabular}$\uparrow$ & \begin{tabular}[c]{@{}c@{}}Motion \\ Smooth\end{tabular}$\uparrow $\\ \hline
Gen-L-Video~\cite{gen-l-video}     & 38.15          & 29.57         & \textbf{98.86\%}         \\
FreeNoise~\cite{freenoise}    & 33.56          & \underline{32.34}        & 97.48\%                                                       \\
FIFO~\cite{fifo}    & 41.25   & 32.15       & 96.83\%                                                       \\
DiTCtrl~\cite{ditctrl}    & \underline{31.64}    & 32.13    & 97.89\%                                                       \\ \hline
Ours     & \textbf{29.89} & \textbf{32.43}     &\underline{98.04\%}                                                   \\ \bottomrule
\label{tab:single}
\end{tabular}
\vspace{-2em}
\end{table}



\subsection{Limitations}
\begin{figure}[h]
\centering
\includegraphics[width=\columnwidth]{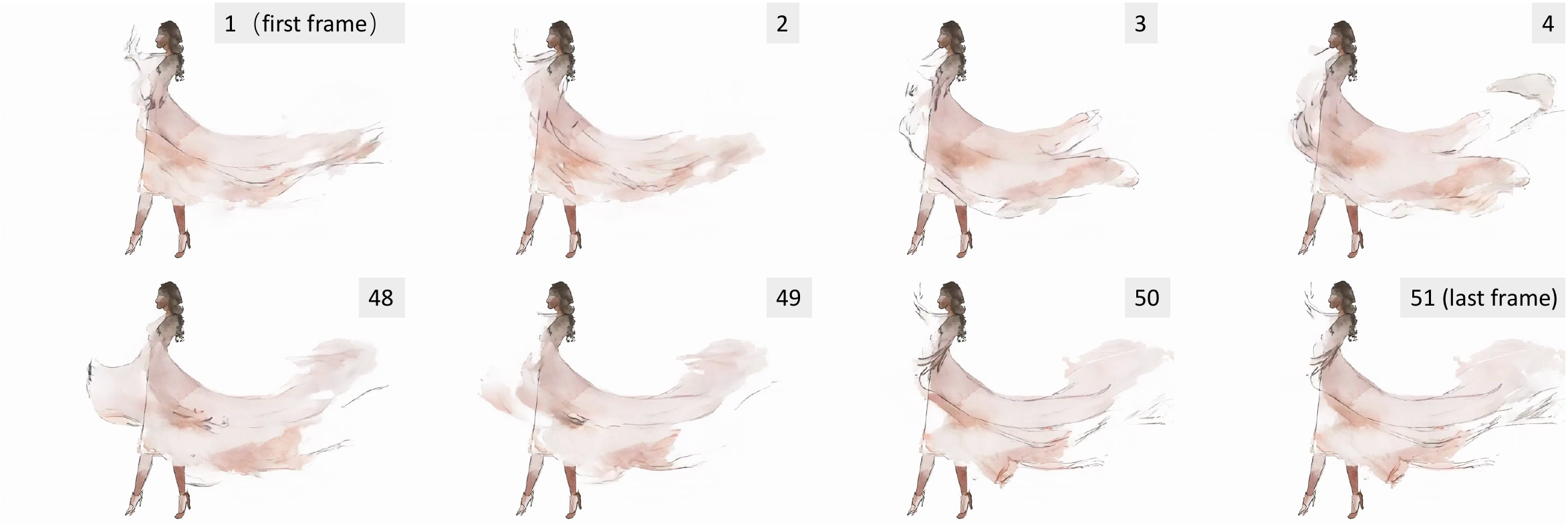}
\vspace{-1em}
\caption{\textbf{Limitation}. The generated results might not show a very smooth video in the customized domain, \eg, the illustration, restricted by the pre-trained text-to-video diffusion model.  }
\label{fig:limitation}
\end{figure}

Since our method is a training-free method based on the pre-trained video diffusion model, our motion prior might be influenced by the pre-trained video diffusion model. As shown in Fig.~\ref{fig:limitation}, we give the results of the successive frame of the generated illustration video. However, the generated dress might not be consistent in the generated results and does not show obvious movement. We argue that this is because of the issues of the motion prior in the pre-trained video diffusion model we use. A better latent diffusion model~\cite{hunyuanvideo, sora} might work better.

\section{Conclusion}
We represent a novel and innovative approach to generating seamlessly looping videos directly from text descriptions without the need for user annotations. This is achieved by repurposing a pre-trained text-to-video latent diffusion model with inference latent modification. In detail, considering each frame should be considered equally in the looping video, we construct a latent cycle and design latent shift to utilize the abilities of the video diffusion model's multi-frame latent denoising in each step, which further expands the scope of seamless looping video generation beyond the limitations of the video diffusion model's context. Besides, we introduce the frame-invariant latent decoding and RoPE-interpolation to further increase the performance.
Compared to previous cinemagraphs, Mobius has a distinct advantage as it does not rely on an image for appearance, thus allowing for more dynamic motion and enhanced visual quality in the generated videos. Through multiple experiments and comparisons, the effectiveness of this method has been verified across different scenarios even on the application of longer video generation task. 

\bibliographystyle{ACM-Reference-Format}
\bibliography{sample-bibliography}

\begin{figure*}[t]
\centering
\includegraphics[width=\textwidth]{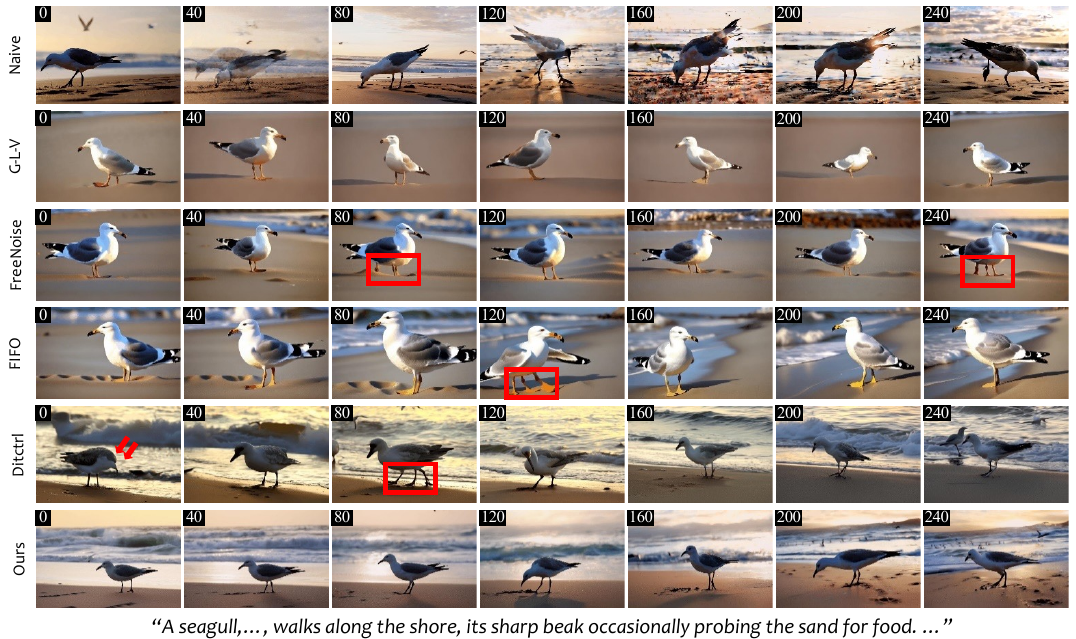}
\vspace{-2em}
\caption{\textbf{Applications on Longer Video Generation}. We show some sampled frames here and the whole video is included in the supplementary video. }
\label{fig:longer}
\end{figure*}

\appendix

\end{document}